\documentclass{article}

\usepackage{arxiv}

\usepackage[utf8]{inputenc} 
\usepackage[T1]{fontenc}    
\usepackage{hyperref}       
\usepackage{url}            
\usepackage{booktabs}       
\usepackage{amsfonts}       
\usepackage{nicefrac}       
\usepackage{microtype}      
\usepackage{graphicx}
\usepackage[utf8]{inputenc}
\usepackage{amsmath}
\usepackage{amsfonts}
\usepackage{amssymb}
\usepackage{todonotes}
\usepackage{dsfont}
\usetikzlibrary{calc}
\usepackage{standalone}
\usepackage{amsthm}

\usepackage{dsfont}

\newtheorem{prop}{Proposition}
\newtheorem{corollary}{Corollary}

\newcommand{\R}{\mathbb{R}}
\newcommand{\N}{\mathbb{N}}

\newcommand{\E}{\mathbb{E}}
\newcommand{\Var}{\mathbb{V}ar}
\newcommand{\Nor}{\mathcal{N}}
\newcommand{\bSigma}{\boldsymbol{\Sigma}}
\newcommand{\bhSigma}{\boldsymbol{\hat{\Sigma}}}
\newcommand{\bsigma}{\boldsymbol{\sigma}}
\newcommand{\bmu}{\boldsymbol{\mu}}

\newcommand{\btheta}{\boldsymbol{\theta}}

\newcommand{\bXi}{\boldsymbol{\Xi}}

\newcommand{\bphi}{\boldsymbol{\phi}}

\newcommand{\bS}{\boldsymbol{S}}
\newcommand{\bzero}{\boldsymbol{\vec{0}}}
\newcommand{\bO}{\boldsymbol{0}}
\newcommand{\bI}{\boldsymbol{I}}
\newcommand{\ELBO}{\mathcal{L}}
\newcommand{\logl}{\boldsymbol\ell}
\newcommand{\hELBO}{\widehat{\mathcal{L}}}
\newcommand{\KL}{D_{KL}}

\newcommand{\Ind}{\mathds{1}}

\newcommand{\ix}{x^{(i)}}
\newcommand{\iy}{y^{(i)}}

\newcommand{\bmuzi}{\bmu_z^{(i)}}

\newcommand{\bhmuzi}{\hat{\bmu}_z^{(i)}}

\newcommand{\bSigmazi}{\bSigma_z^{(i)}}
\newcommand{\bhSigmazi}{\bhSigma_z^{(i)}}

\newcommand{\bSzi}{\bS_z^{(i)}}
\newcommand{\tr}{\textnormal{tr}}
\newcommand{\diag}{\textnormal{diag}}
\newcommand{\Wj}{W_{j\cdot}}

\usepackage{subfigure}
\usepackage{makecell}

\usepackage{authblk}

\usepackage[symbol]{footmisc}

\title{A lower bound for the ELBO of the Bernoulli Variational Autoencoder}

\author[1]{Robert Sicks\thanks{Corresponding author}}
\author[2]{Ralf Korn}
\author[1]{Stefanie Schwaar}
\affil[1]{Department of Financial Mathematics, Fraunhofer ITWM, Kaiserslautern, Germany}
\affil[2]{Department of Financial Mathematics,	Technical University of Kaiserslautern, Kaiserslautern, Germany}
\renewcommand{\headeright}{Technical Report}
\renewcommand{\undertitle}{Technical Report}

\begin{document}
\maketitle

\begin{abstract}
We consider a variational autoencoder (VAE) for binary data. Our main innovations are an interpretable lower bound for its training objective, a modified initialization and architecture of such a VAE that leads to faster training, and a decision support for finding the appropriate dimension of the latent space via using a PCA. Numerical examples illustrate our theoretical result and the performance of the new architecture. 
\end{abstract}


\section{Introduction}
\label{sec:Introduction}
With the Variational Autoencoder (VAE), \cite{Kingma} present a special form of autoencoder that incorporates randomness to the architecture. This form of autoencoder yields an interpretable latent space when it comes to images as input. \cite{Goodfellow2016} describe VAE as an ``excellent manifold learning algorithm'' due to the fact that the model is forced ``to learn a predictable coordinate system that the encoder can capture''. Furthermore, VAE are suitable for generating new unobserved data by interpolating within the latent space (see the appendix A in \cite{Kingma} for vivid examples of moving through the latent space). 

In this paper we analyse VAE for the case of binary data. Let $x^{(1)},\ldots, x^{(N)} \in \{0,1\}^d$ be such observations. A VAE consists of an encoder and a decoder. The encoder maps these multivariate binary observations to a low-dimensional numeric space of dimension $\kappa \ll d$. The decoder applies the inverse operation. \cite{Kingma} introduce the ``Bernoulli MLP decoder'' for the binary setting used here. We call a VAE with this type of decoder ``Bernoulli VAE''.

This work contributes to the understanding of VAE in three areas:
\begin{enumerate}
	\item We derive an interpretable lower bound for the training objective of a Bernoulli VAE (the ELBO). 
	\item We propose an initialization scheme with an architecture change for Bernoulli VAE that results in faster convergence.
	\item We derive a decision support for finding the parameter of the latent space dimension $\kappa$ with a principal component analysis (PCA).
\end{enumerate}

Probably the most comparable work in this research field is \cite{Lee2010}. They present a sparse logistic PCA, which has a target objective  similar to the Bernoulli VAE: A balance between the reconstruction of the binary inputs and a regularization term. \cite{Kunin2019} show that regularization helps the linear autoencoder to learn principal directions. Without regularization the linear autoencoder learns the subspace spanned by the top principal direction but not the directions themselves.
\cite{Dai2018} provide an interpretable term for the ELBO of a VAE with a ``Gaussian MLP decoder'' (as denoted in \cite{Kingma}). This term is similar to our bound. \cite{Tipping1999} derive an equivalent term, again under Gaussian assumptions but from the perspective of a PCA instead of a VAE. 

For this paper, we assume $X_1,\ldots,X_d$ given $Z$ to be mutually independent and Bernoulli distributed. This assumption yields what is known as the ``Cross Entropy Loss''. This loss finds application in different fields:
\cite{Blaschke2018} apply VAE on molecules represented in a so-called ``SMILE''-format. \cite{Germain2015} develop a modification for autoencoders to consider autoregressive constraints and focus on binary inputs.
\cite{Duan2019} apply the Bernoulli VAE to images.\footnote{Even though images are usually not represented in binary fashion, the pixel range can be transformed to be within zero and one. E.g. if the values originally range from 0 to 255 we divide them by 255.}

This article is structured as follows: In section \ref{sec:Background}, we present related work and compare our mathematical problem in detail to the existing literature. Theoretical findings and discussion can be found in section \ref{sec:A lower bound for the ELBO}. In section \ref{sec:Simulation}, we look at our simulation setup and present our results. Finally, we summarize in section  \ref{sec:Conclusion and possible extensions} our contribution to the research of VAE and highlight further research topics.

\section{Background and related work}
\label{sec:Background}
The Bernoulli VAE yields a continuous representation of binary data, which can be viewed as a kind of logistic PCA. In this chapter, we have a look at related work and compare the mathematical formulations of sparse logistic PCA and Bernoulli VAE.

\subsection{Related work: Logistic PCA}
\cite{Tipping1999} view PCA as a maximum likelihood procedure and introduce it as a probabilistic PCA (pPCA). We show connections of the Bernoulli VAE to pPCA based on our bound. \cite{Collins2002} generalize pPCA and derive an optimization for the exponential family. They provide a generic algorithm for minimizing the loss. \cite{DeLeeuw2006} presents logistic (and probit) forms of PCA and a majorization algorithm to fit the parameters. \cite{Lee2010} propose a sparse logistic PCA by introducing $L_1$ regularization. This can be seen as the deterministic version of our proposed method here as we will see in section \ref{subsec:Comparison of sparse logistic PCA and Bernoulli VAE}. \cite{Landgraf2015} formulate the logistic PCA differently: For binary data their model is not based on matrix-factorization but on projection of the data.

\subsection{Related work: Autoencoder}

\cite{Bourlard1988} show for autoencoders with linearizable activations that the optimal solution is given by the solution of a Singular Value Decomposition. \cite{Baldi1989a} extend these results and analyse the squared error loss of autoencoders for all critical points. \cite{Saxe2014} provide theory for learning deep linear neural networks for different non-linear dynamics. \cite{Josse2016} present an autoencoder that provides a stable encoding when perturbed with bootstrap noise. Therefore, it can adapt to non-isotropic noise structures. \cite{Pretorius2018} analyse the learning dynamics of linear denoising autoencoders. \cite{Dai2018} analyse the loss of the Gaussian VAE. They show connections to pPCA and robust PCA. The bound of the ELBO from this paper has an equivalent form to one of their results. \cite{Kunin2019} consider regularizations in linear autoencoders and analyse the critical points for three targets.

VAE experience regularization over the Kullback-Leibler-Divergence (KL-Divergence) term (see below). Variations of autoencoders exists, which regularize the closeness of distributions in a different way. Adversarial Autoencoders (AAE) (see \cite{Makhzani2015}) try to approximate the Jensen-Shannon-Divergence instead of using the KL-Divergence. \cite{Zhao2019} motivate a unifying model for VAE and AAE.

\subsection{Comparison of sparse logistic PCA and Bernoulli VAE}
\label{subsec:Comparison of sparse logistic PCA and Bernoulli VAE}

We are interested in a sparse low-rank representation of binary data. One possible approach is the logistic PCA introduced by \cite{Lee2010}. Let $x^{(1)},\ldots, x^{(N)} \in \{0,1\}^d$ be the observations of $X$, a multivariate Bernoulli distributed random variable, having a low-rank representation. Let $(a^{(1)},\ldots, a^{(N)})^T = A\in \R^{N\times \kappa}$ denote these representations with $\kappa < < d$, i.e. there exists $ (b_1| \ldots| b_\kappa) = B \in \R^{d \times \kappa}$ and a location vector $\vec{b_0} \in \R^d$ such that
\[
P(X = x^{(i)}) \sim Bern\left(p(\vec{b_0},a^{(i)},B)\right),
\]
where $p(\vec{b_0},a^{(i)},B)\in [0,1]^d$ denotes the probability vector.
Finding the sparse low-rank representation (in the spirit of \cite{Lee2010}) leads to maximizing the log-likelihood and a penalization term, given by
\begin{equation}\label{eq: log-Lee et al}
\logl(\vec{b_0},A,B) - N \cdot P_\lambda(B).
\end{equation}

\cite{Lee2010} introduced a $L_1$ penalty term to get sparse loading vectors $b_1, \ldots, b_\kappa$, similar to the LASSO regression from \cite{Tibshirani1996}. 

We are going to analyse an alternative to logistic PCA using the VAE introduced by \cite{Kingma}. For the VAE setup, we add a latent variable $Z$ with density $q_{\bphi}(z | X)$ to this model. We change our perspective from 
\[
P(X = x^{(i)}) \sim Bern\left(p(\vec{b_0},a^{(i)},B)\right),
\]
to 
\[
P_{\btheta}(X|Z) \sim Bern\left(p_{\btheta}\left(Z\right)\right),
\]
where the probability vector $p_{\btheta}(Z) \in [0,1]^d$ is a function of $Z$ and the model parameters $\btheta$.
We interpret this as replacing the deterministic low-rank representation $a^{(i)}$ by a stochastic $z^{(i)}$.

Using the Bayes theorem, we get
\begin{align*}
\ln P_{\btheta}(\ix) \geq & \; \E_{q_{\bphi}(\cdot|\ix)}\left[\log P_{\btheta}(\ix|Z)\right] - \KL\left(q_{\bphi}(Z|\ix)|| p(Z)\right), 
\end{align*}
which gives a lower bound for the log-likelihood function: The ``Evidence Lower Bound'' (ELBO). 
Since in our current model formulation the log-likelihood is intractable, we maximize the ELBO instead and assume the corresponding log-likelihood to increase as well.

The first part $E_{q_{\bphi}(\cdot|x)}[\log P_{\btheta}(x|Z)]$ resembles the reconstruction error of an autoencoder: Given a sample $z\sim q_{\bphi}(\cdot|x)$, we integrate over the probabilities that an observation x was generated from that sample. The expression $q_{\bphi}(\cdot|x)$ is the generating function that resembles the encoder in the autoencoder. $P_{\btheta}(x|Z)$ can be identified with the decoder of an autoencoder. Both, encoder and decoder are usually implemented as a multi layer perceptron (MLP).
The second term is the Kullback-Leibler-Divergence that indicates how close the two inputs are. The further these two expressions are apart, the higher the penalization.
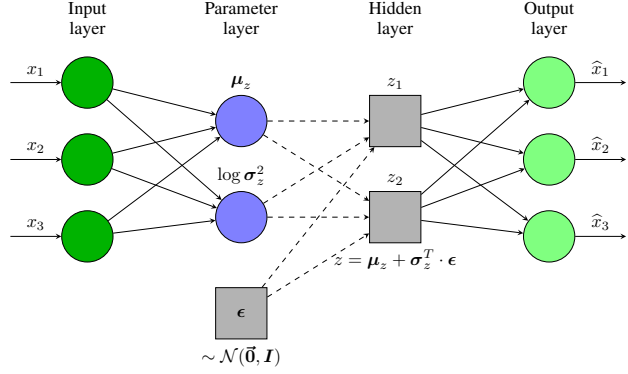
\begin{figure}
	\centering
		\tikzset{%
		output neuron/.style={
			circle,
			draw,
			minimum size=1cm,
			fill=green!50
		},
		hidden neuron/.style={
			circle,
			draw,
			minimum size=1cm,
			fill=blue!50
		},
		input neuron/.style={
			circle,
			draw,
			minimum size=1cm,
			fill=black!30!green
		},
		input epsilon/.style={
		rectangle,
		draw,
		minimum size=1cm,
		fill=black!30
		}	
	}
	
	\begin{tikzpicture}[x=1.5cm, y=1.5cm, >=stealth]
	
	\foreach \m/\l [count=\y] in {1,2,3}
	\node [input neuron/.try, neuron \m/.try] (input-\m) at (0,2.5-\y) {};
	
	\foreach \m [count=\y] in {1,2}
	\node [hidden neuron/.try, neuron \m/.try ] (param-\m) at (2,2.25-\y*1.25) {};
	
	\node [input epsilon/.try, neuron 3/.try ] (param-3) at (2,2.25-3.75) {};
	
	\foreach \m [count=\y] in {1,2}
	\node [input epsilon/.try, neuron \m/.try ] (hidden-\m) at (4,2.25-\y*1.25) {};
	
	\foreach \m [count=\y] in {1,2,3}
	\node [output neuron/.try, neuron \m/.try ] (output-\m) at (6,2.5-\y) {};
	
	\foreach \l [count=\i] in {1,2,3}
	\draw [<-] (input-\i) -- ++(-1,0)
	node [above, midway] {$x_\l$};
	
	\foreach \l [count=\i] in {1,2}
	\node [above] at (hidden-\i.north) {$z_\l$};
	
	\node [above] at (param-1.north) {$\boldsymbol{\mu}_z$};
	\node [above] at (param-2.north) {$\log \boldsymbol{\sigma}^2_z$};
	\node at (param-3) {$\boldsymbol{\epsilon}$};
	\node [below] at (param-3.south) {$ \sim \mathcal{N}(\boldsymbol{\vec{0}},\boldsymbol{I})$};

	\node [below] at (hidden-2.south) {$z=\boldsymbol{\mu}_z + \boldsymbol{\sigma}_z^T \cdot \boldsymbol{\epsilon}$};
	
	\foreach \l [count=\i] in {1,2,3}
	\draw [->] (output-\i) -- ++(1,0)
	node [above, midway] {$\widehat{x}_\l$};
	
	\foreach \i in {1,...,3}
	\foreach \j in {1,...,2}
	\draw[->] (input-\i) -- (param-\j);

	\foreach \i in {1,...,3}
	\foreach \j in {1,...,2}
	\draw [dashed,->] (param-\i) -- (hidden-\j);
	
	\foreach \i in {1,...,2}
	\foreach \j in {1,...,3}
	\draw [->] (hidden-\i) -- (output-\j);

	\foreach \l [count=\x from 0] in {Input, Parameter, Hidden, Output}
	\node [align=center, above] at (\x*2,2) {\l \\ layer};
	
	\end{tikzpicture}
	\caption{The implementation of a VAE as a neural net. The outputs on the right side should resemble the inputs on the left as good as possible. The solid arrows in the picture stand for the biases and  weights. The dashed arrows correspond to the calculation procedure $z=\bmu_z + \bsigma_z^T \cdot \epsilon$. The most important feature of a VAE compared to a normal autoencoder is the stochastic source, from which we sample to get an output. Note that ``$\bmu_z$'' and ``$\log\bsigma_z^2$'' are $\kappa$-dimensional and (for this picture) we assume an independent Gaussian distribution of $Z$.}\label{fig:tikz_VAE}
\end{figure}
Figure \ref{fig:tikz_VAE} shows the implementation of a VAE as a neural net. 

There are several similarities to the work of \cite{Lee2010}:
\begin{itemize}
	\item The ELBO and the loss in \eqref{eq: log-Lee et al} consist of two parts: The first part provides information on the goodness of fit, while the second part is a penalty term.
	\item To minimize the loss in \eqref{eq: log-Lee et al} \cite{Lee2010} use a ``Majorization-Minimization'', algorithm based on a second order Taylor series expansion. For the proof of proposition \ref{wohoo-i-got-a-proposition}, we also use a second order Taylor series expansion.
	\item The probability vector of \cite{Lee2010} $p(\vec{b_0},a^{(i)},B)$ and our assumed probability vector $p_{\btheta}(Z)$ use sigmoid activation.
\end{itemize}

\section{Theoretical results}
\label{sec:A lower bound for the ELBO}

In this section, we present our theoretical findings. Recall that 
\[
\ln P_{\btheta}(\ix) \geq ELBO.
\] 
Our central result is the lower bound for the ELBO in proposition \ref{wohoo-i-got-a-proposition}. So instead of the ELBO, we maximize the lower bound and assume that the corresponding log-likelihood increases.

We first state our setting and assumptions. Then we formulate proposition \ref{wohoo-i-got-a-proposition} and interpret the resulting lower bound of the ELBO. Based on these findings, we introduce an initialization and a change of architecture for Bernoulli VAE. Finally, we propose a way to determine the latent space dimension $\kappa$ given the training data.

\subsection{Assumptions and lower bound}\label{subsec:Assumptions and lower bound}

Let $x^{(1)}, \ldots, x^{(N)} \in \left\{0,1\right\}^{d}$ denote the observations, where $d \in \N$ denotes the dimension. We assume a latent variable model: The values $x^{(1)}, \ldots, x^{(N)}$ were generated given a latent (not observable) sample $z$ of a $\kappa$-dimensional variable $Z$. Furthermore, we assume $d\gg\kappa$ and $X_1,\ldots,X_d$ given $Z$ are mutually independent. The distribution of $X$ given $Z$ is 
\begin{equation}\label{eq:X given Z Bern}
P_{\btheta}(X|Z) \sim Bern\left(p_{\btheta}\left(Z\right)\right),
\end{equation}
where the probability vector $p_{\btheta}(Z) \in [0,1]^d$ is a function of $Z$ and the model parameters $\btheta$. We fix the probability vector $p_{\btheta}\left(Z\right)$ the following way:
For each entry of $p_{\btheta}(z) \in [0,1]^d$, we have
\[
p_{\btheta,j}(Z):= \sigma([WZ+b]_j) = \dfrac{1}{1  + e^{-[WZ+b]_j}}
\]
with $W \in \R^{d\times \kappa}$, $b\in \R^d$ and $[\;\cdot\;]_j$ the $j$-th entry. $\sigma( \cdot)$ is the sigmoid function. Hence $\btheta := \{W,b\}$ in our case. $W$ and $b$ can be identified with the weights and biases of the decoder of the VAE. In figure \ref{fig:tikz_VAE}, these are the arrows between ``Hidden Layer'' and ``Output Layer''.

We further expect the distribution of the variables $Z$ to be given by the standard recognition model (see \cite{Kingma}): 
\[
q_{\bphi}(z|X) \sim \Nor(\bmu_z,\bSigma_z),
\]
where we define $\bmu_z$ and $\bSigma_z$ as functions of $X$ and $\bphi$. Furthermore, we define $p(z)$ as the density of a standard normal distribution $\Nor(\bzero, \bI)$.

The target of a VAE is the average ELBO, given by
\begin{align}
\ELBO(\bphi,\btheta) :=& \dfrac{1}{N} \sum_{i=1}^N \bigg(\E_{Z \sim q_{\bphi}\left(\cdot|\ix\right)}\left[\log P_{\btheta}(\ix|Z)\right] -\KL\left(q_{\bphi}(Z|\ix)|| P(Z)\right) \bigg),\label{eq:bin-ELBO_up_here}
\end{align}
for which we state the following lower bound:

\begin{prop}\label{wohoo-i-got-a-proposition}
	Given the assumptions stated above, we further assume that for the encoder we have
	\begin{itemize}
		\item $\bmuzi := f_1(\ix,\bphi)$ and
		\item $0 \prec \bSigmazi := \bSzi{\bSzi}^T$, with $\bSzi := f_2(\ix,\bphi),$
	\end{itemize}
	where $f_1$ and $f_2$ are arbitrary functions that include affine transformations. Then, there exists a lower bound for the average ELBO in \eqref{eq:bin-ELBO_up_here} that admits an optimal solution for $\bmuzi$ and $\bSigmazi$ such that it can be written as
	\begin{align}
	\hELBO(W,b) = \dfrac{1}{N}\sum_{i=1}^N \Bigg[& -\dfrac{1}{2} \Big(4 \iy - b\Big)^T C^{-1} \Big( 4 \iy - b\Big) \Bigg] - \dfrac{1}{2}\log|C| + \dfrac{d}{2}, \label{eq:prop_lowerBound}
	\end{align}
	where $\iy:=\ix - \dfrac{1}{2}\Ind\in \bigg\{-\dfrac{1}{2},\dfrac{1}{2}\bigg\}^d$ with $\Ind:=(1,\ldots,1)^T \in \R^d$ and $C:=(4I_{d} + WW^T)$. 
\end{prop}

The proof of this result can be found in the supplementary material. The advantage of this lower bound is the gain in interpretability as we will see. The assumptions of proposition \ref{wohoo-i-got-a-proposition} for $\bmuzi$ and $\bSigmazi$ can be interpreted as ``the encoder is always doing its job''. These are the same as in lemma 1 of \cite{Dai2018}.

The resulting objective is equivalent to the result of lemma 1 in \cite{Dai2018} and the log-likelihood objective in \cite{Tipping1999}.  Contrary to our Bernoulli distribution assumption here, their objective originated from a Gaussian distribution. Nonetheless both assume a latent variable model, as we do here. The properties derived by \cite{Dai2018} and \cite{Tipping1999} apply to our target \eqref{eq:prop_lowerBound}. Therefore, if we assume $\bSigma_z$ to be diagonal (which is a standard assumption for VAE) we get an upper bound for \eqref{eq:prop_lowerBound} (see \cite{Dai2018}).

In the following we look at the optimal values of $W$ and $b$  for \eqref{eq:prop_lowerBound} and highlight important properties. To do so, we rewrite \eqref{eq:prop_lowerBound} in the form of the objective in \cite{Tipping1999}:
\[
\hELBO(W,b) = \dfrac{1}{2} \left(d - \ln|C| - \tr\left(C^{-1}S\right)\right),
\]
with
\begin{equation}\label{eq:S}
S:= \dfrac{1}{N}\sum\limits_{i=1}^{N} \left(4\iy - b\right)\left(4\iy - b\right)^T
\end{equation}
the sample covariance matrix (of $4\iy$), since according to \cite{Tipping1999} the maximum point for $b$ is given by the sample mean (of $4\iy$), as follows:
\begin{equation}\label{eq:opt_b}
\hat{b} =4 \bar{y} :=  \dfrac{4}{N} \sum_{i=1}^{N}\iy.
\end{equation}
The maximum point for $\hELBO$ w.r.t. $W$ is given by
\begin{equation}\label{eq:opt_w}
\hat{W}= U_{\kappa} (K_{\kappa}- 4 I_{\kappa})^{1/2} R,
\end{equation}
where $U_{\kappa} \in \R^{d\times\kappa}$ is composed of $\kappa$ eigenvectors of the matrix S. These eigenvectors are associated with the $\kappa$ biggest eigenvalues $\lambda_1, \ldots,\lambda_\kappa$. $K_{\kappa} \in \R^{\kappa\times\kappa}$ is a diagonal matrix with entries 
\begin{equation}\label{eq: k bigger 4}
k_j=\left\{\begin{array}{ll} \lambda_j, & \lambda_j \geq 4 \\
4, & \textnormal{else}\end{array}\right. .
\end{equation}
$R \in \R^{\kappa\times\kappa}$ is an arbitrary rotation matrix. \cite{Kunin2019} obtain similar optimal values for the regularized linear autoencoders.
We can derive several interesting features of $\hat{W}$:
\begin{enumerate}
	\item It is possible to have $\textnormal{rank}(\hat{W})< \kappa$. We use this observation in section \ref{subsec: Optimal latent space size} for the choice of the latent space size of the VAE.
	\item The rotation matrix $R$ is arbitrary. This implies that our optimal solution is invariant to rotations. \cite{Dai2018} show this as well as invariance to permutations in their theorem 2.
\end{enumerate}

It can be shown that other candidate points of $\hELBO(W,b)$ for $W$ represent saddle points (see appendix of \cite{Tipping1999}). So, if we think of training a VAE, the optimizer should be robust to saddle points. 

It is important that we assume $X_1,\ldots,X_d$ given $Z$ are mutually independent with \eqref{eq:X given Z Bern} and not that the $X_1,\ldots,X_d$ are mutually independent with 
\begin{equation}\label{eq:X Bern}
X \sim Bern\left(p_{\btheta}\right).
\end{equation}
S from \eqref{eq:S} with b as in \eqref{eq:opt_b} is the sample covariance matrix for the variable $4Y:= 4\left(X-1/2 \Ind\right)$. If we take the expectation under assumption \eqref{eq:X Bern}, with $X_1,\ldots,X_d$ mutually independent, we get
\begin{align*}
\E\left(S\right) &= 16 \cdot \diag(\Var(X_1),\ldots, \Var(X_d))&\\
&= 16 \cdot \diag\left(p_{\btheta,1}\cdot\left(1-p_{\btheta,1}\right),\ldots, p_{\btheta,d}\cdot\left(1-p_{\btheta,d}\right)\right).
\end{align*}
So, we have all the eigenvalues $\lambda_1, \ldots,\lambda_d$ of $\E\left(S\right)$ readily at hand and $\lambda_j$ only depends on $p_{\btheta,j} \in [0,1]$ for $j=1,\ldots,d$. The inequality in \eqref{eq: k bigger 4} is only fulfilled at $p_{\btheta,j}=0.5$, which results in $\hat{W} = \bO$. Thus assumption \eqref{eq:X Bern}, with $X_1,\ldots,X_d$ mutually independent, does not lead to meaningful solutions for the weights of the decoder.\\
Under assumption \eqref{eq:X given Z Bern}, with $X_1,\ldots,X_d$ given $Z$ mutually independent, we cannot calculate $\E(S)$ as above: $X_1,\ldots,X_d$ are not mutually independent without knowing $Z$. This fact stems from the common parent argument in Bayesian network theory. \\
We generate data in \ref{subsec:Data generation} according to assumption \eqref{eq:X given Z Bern}. For this data we observe eigenvalues that are bigger than the needed value of $4$.\footnote{E.g.: $8.44$, $5.96$, $5.42$, ... .}

Apart from the optimal $W$ and $b$ we achieve representations for optimal $\bmuzi$ and $\bSigmazi$:
\begin{corollary}\label{cor:mu,Sigma}
	For an observation $\ix \in \{0,1\}^d$ the optimal closed form solutions for $\bmuzi$ and $\bSigmazi$ of the lower bound of the ELBO are given by
	\begin{equation}\label{corr:opt-bSigmaz}
	\bhSigma_z := \bhSigmazi = (I_{\kappa} + \dfrac{1}{4} W^TW)^{-1},
	\end{equation}
	and
	\begin{align}
	\bhmuzi &= \bhSigma_z W^T\Big(\ix - \dfrac{1}{2} \Ind- \dfrac{1}{4} b\Big) \nonumber \\
	&= \bhSigma_z W^T\Big(\iy - \dfrac{1}{4} b\Big).\label{corr:opt-bmuzi}
	\end{align}
\end{corollary}

These optimal solutions are a by-product of the proof of proposition \ref{wohoo-i-got-a-proposition}. They give us insight into what to expect from a trained neural net. We use this to propose a new initialization and a change of the VAE structure.

\subsection{Initialization and change of net architecture for Bernoulli VAE}
\label{subsec:Initialization for Variational Autoencoder}

Given the optimal values for the encoder outputs in corollary \ref{cor:mu,Sigma} as well as for the decoder parameters in \eqref{eq:opt_w} and \eqref{eq:opt_b}, we initialize the corresponding weights and biases of a VAE. This affects only the last layers of the encoder and decoder.
Then we introduce and discuss a slight change of the net architecture.
At last we explain our approach for over-parametrized nets in the form that we have more ingoing dimensions into the last layers of encoder and decoder than needed. This issue arises when we consider hidden layers with higher dimension than the input.\\
Our simulation results in \ref{subsec:Training setting and results} show a faster convergence and comparable results as produced by a VAE without this change.

The weights and biases for the last decoder layer are straightforward set as the optimal $\hat{W}$ from \eqref{eq:opt_w} and $\hat{b}$ from \eqref{eq:opt_b}. 
For the last encoder layers we first note that there are two layers: the ``$\bmu_z$''- and the ``$\log\bsigma_z^2$''-layer. We initialize the weights and biases of the ``$\bmu_z$''-layer as follows:
\begin{itemize}
	\item $\hat{W}_e := \hat{\bSigma}_z\hat{W}^T$.
	\item $\hat{b}_e := \hat{\bSigma}_z\hat{W}^T \left(- \dfrac{1}{2} \Ind- \dfrac{1}{4} \hat{b}\right).$
\end{itemize}

The VAE-architecture is changed at the ``$\log\bsigma_z^2$''-layer: We decouple it from the layers before. Consequently, the output of this layer never changes. We set it to produce the $\log$ of $\diag(\hat{\bSigma}_z)$. This change is justified by the fact that $\hat{\bSigma}_z$ in \eqref{corr:opt-bSigmaz} is the same for all inputs $\ix$.
Apart from a better initialization than over random values, a benefit of this approach is that the $Z$ distribution now is prohibited to become degenerated. E.g. is this the case when $\bSigma_z \rightarrow \bO$.

All weights and biases produced here only depend on $\hat{W}$ and $\hat{b}$, which are easily obtainable: We just need to calculate the sample mean and a Singular Value Decomposition of the sample covariance.

In case of over-parametrized nets, more edges lead into the affected layers than we need. This problem only concerns the weights and not the biases. We solve this by initializing not needed dimensions of the weights with zero. This still allows the net to change these weights during training, but they have a starting point which is optimal in the sense of our bound from proposition \ref{wohoo-i-got-a-proposition}.

\subsection{Optimal latent space size}\label{subsec: Optimal latent space size}

Given the structure of $\hat{W}$ in \eqref{eq:opt_w}, we propose a decision support for the choice of the latent space dimension $\kappa$. The approach we propose here is based on PCA and is therefore not new. Usually we perform a Singular Value Decomposition of the estimated covariance matrix and consider the dimensions associated to the biggest eigenvalues. This way, we retain the variation present in the data up to a maximum amount. The parameter $\kappa$ equals the number of eigenvalues we decide for.

The form of \eqref{eq:opt_w} shows which eigenvalues of the sample covariance matrix S are eligible for $\hat{W}$: Only those that are greater then $4$. Hence, all dimensions with smaller associated eigenvalues can be discarded as this would only introduce a zero-row in $\hat{W}$.

Concluding, the value for $\kappa$ can be chosen the following way:
\begin{enumerate}
	\item Create $S$ as in \eqref{eq:S} with b as in \eqref{eq:opt_b}.
	\item Perform a Singular Value Decomposition of S.
	\item Only consider eigenvalues as eligible that are greater than $4$ when choosing $\kappa$
\end{enumerate}

For the equivalent terms in \cite{Tipping1999}, \cite{Dai2018} and \cite{Kunin2019}, this approach is not directly possible. There, $\hat{W}$ also depends on the parameter $\bsigma_x^2$, the variance of the observation distribution under isotropic Gaussian assumption. This parameter has to be estimated.

\section{Simulation}\label{sec:Simulation}

Given our theoretical results, we provide simulations that back up these findings. Therefore, in this section, we compare performances of VAEs.
The essential messages of our simulations are the following:
\begin{enumerate}
	\item Given a VAE, our proposed initialization of the net results in a faster training convergence.
	\item Even without our initialization as starting points, the VAE approaches the theoretical bound after a longer training period.
	\item With our initialization and architecture, the VAE is less prone to over-fitting.
\end{enumerate}

For the simulations, we generate data and show that our theory applies to the general case. Practitioners are welcome to apply our proposed setting to their data.

We first describe the data generation in detail. Then, we show the architecture and the initialization of the VAE and how we train the two resulting nets. Given these two parts, we present the simulation setup and interpret the results.

\subsection{Data generation}\label{subsec:Data generation}

For $k=2$, $N \in \{100;5000;10000\}$ and $d \in \{200;400;1000\}$ we generate two matrices $A \in \R^{N\times k}$ and $B \in \R^{d \times k}$. $A$ is identifiable  with the principal components and B with the loading vectors of a PCA. $B$ is assumed to be sparse (see below).

We construct the matrices in the fashion of \cite{Lee2010}: The two-dimensional principal components $a^{(i)} (i=1,\ldots,N)$ of $A$ are drawn from normal distributions, so that $a^{(i)}_1 \sim \Nor(0,0.09)$ and  $a^{(i)}_2 \sim \Nor(0,0.25)$. The sparse loading vectors are constructed by setting $B$ to zero except for $b_{j,1} = 1, j =1,\ldots,20$ and $b_{j,2} = 1, j =21,\ldots,40$.

Given $A$ and $B$ we calculate
\[
\bXi := A \cdot B^T
\]
and the probability matrix $\Pi$, with
\[
\Pi = \sigma( \bXi),
\]
where we apply the sigmoid function $\sigma(\cdot)$  element-wise.
We then use the probabilities $\Pi^{(i)}_{j}$ to independently draw samples 
\[
x^{(i)}_{j} \sim Bern(\Pi^{(i)}_{j})
\]
and with the data $X_{Data} := (x^{(i)}_{j})_{i=1,\ldots,N;j=1,\ldots,d}$, we conduct the simulation.

Given the combinations of $N$ and $d$, we get nine different data sets for our simulation.

Though the case $N=100$ seems to be rather odd for training of neural nets,  \cite{Lee2010} generate data explicitly with $N=100$ and $d \in \{200;400;1000\}$ and denote it as challenging tasks for their logistic PCA. We therefore keep it in our simulation study to see how our bound from proposition \ref{wohoo-i-got-a-proposition} relates to the loss of the VAEs for the test data.

\subsection{VAE architecture}
\label{subsec:VAE architecture}

The architecture of the VAE that we look at can be described as follows:
\begin{align*}
x(d)\rightarrow E_1(2000)\rightarrow E_2(1000)&\rightarrow \; \bmu_z(k)\; \rightarrow D_1(1000)\rightarrow D_2(2000)\rightarrow \hat{x}(d),\\
&\searrow\log\bsigma_z^2(k)\nearrow& \\
\end{align*}
where the dots just indicate the line break. Each aspect denotes a layer and the value in the parentheses gives the dimension of this layer. So, $E_1,E_2$ denote the hidden layers of the encoder and $D_1,D_2$ those of the decoder. This architecture, the notation and the training parameters (see next section) are as in \cite{Dai2018}.

The hidden layers of the encoder and the decoder are implemented with ``ReLU''-activation (see \cite{nair2010rectified}), which is known to be highly expressive. The ``$\bmu_z$''- and ``$\log\bsigma_z^2$''- layer have linear activations. As we need an output between zero and one and in consistency with our theoretical derivation from before, the last layer ``$\hat{x}$'' has a sigmoid activation function.

The network weights and biases are initialized as proposed by \cite{He2015}. This initialization particularly considers rectifier non-linearities.

Apart from the fact that we expect the net to provide a better loss than provided by our theoretical bound, the net should also be able to represent the optimal $\bhmuzi$ and diagonal entries of the optimal $\bhSigmazi$ from corollary \ref{cor:mu,Sigma} if necessary.

We train a second version of VAE, called ``VAE preinit''. This is equal to the one described above for all but the following two aspects:

\begin{enumerate}
	\item We initialize the weights and biases of the ``$\bmu_z$''- and ``$\hat{x}$''-layer as proposed in \ref{subsec:Initialization for Variational Autoencoder}. Dimensions that we do not need for this initialization are set to zero but can change during training.
	
	\item We fix the ``$\log\bsigma_z^2$''-layer to always produce the diagonal entries of $\bhSigma_z$ from corollary \ref{cor:mu,Sigma}. The output of this layer is not affected by training. So the variance of ``VAE preinit'' does not depend on the input and behaves approximately as suggested from the theoretical results. 
\end{enumerate}

Obviously both autoencoders are over-parametrized for the simple simulation. Looking at the decoder part and comparing it with the data generation in \ref{subsec:Data generation}, one layer suffices.
Furthermore, our theoretical setting in section \ref{sec:A lower bound for the ELBO} manifests this view. We note that the simulation on such a ``canonical'' version of a VAE produced results that favoured our proposed initialization and architecture even more than those stated in \ref{subsec:Training setting and results}. The results for this case can be found in the supplementary materials.
The intention behind the over-parametrized structures is to show that even for deep learning architectures, our initialization performs well and that our theoretical bound serves its purpose.

\subsection{Training setting and results}
\label{subsec:Training setting and results}

For training of the nets we used the Adam optimizer by \cite{Kingma2015} with learning rate $0.0001$ and a batch size of $100$. Training was done for a total of $400$ epochs each time. We split the data into $2/3$ training data and $1/3$ test data and calculate the optimal values $\hat{W},\hat{b},\hat{\bmu}_z$ and $\hat{\bSigma}_z$ only on basis of the training data. Given these, we obtain a theoretical bound $\hELBO$ for training and test data. We also use these parameters for the initialization of ``VAE preinit'' described in \ref{subsec:VAE architecture}.

We first present our simulation results given the values of the losses after the $400$ epochs. Afterwards, we compare the development of the two autoencoders during one training.

\subsubsection{Simulation results for deep architectures}\label{subsubsec:Simulation results for the deep architectures}

\begin{figure*}[ht]
	\vskip 0.2in
	\begin{center}
		\centerline{\includegraphics[width=\columnwidth]{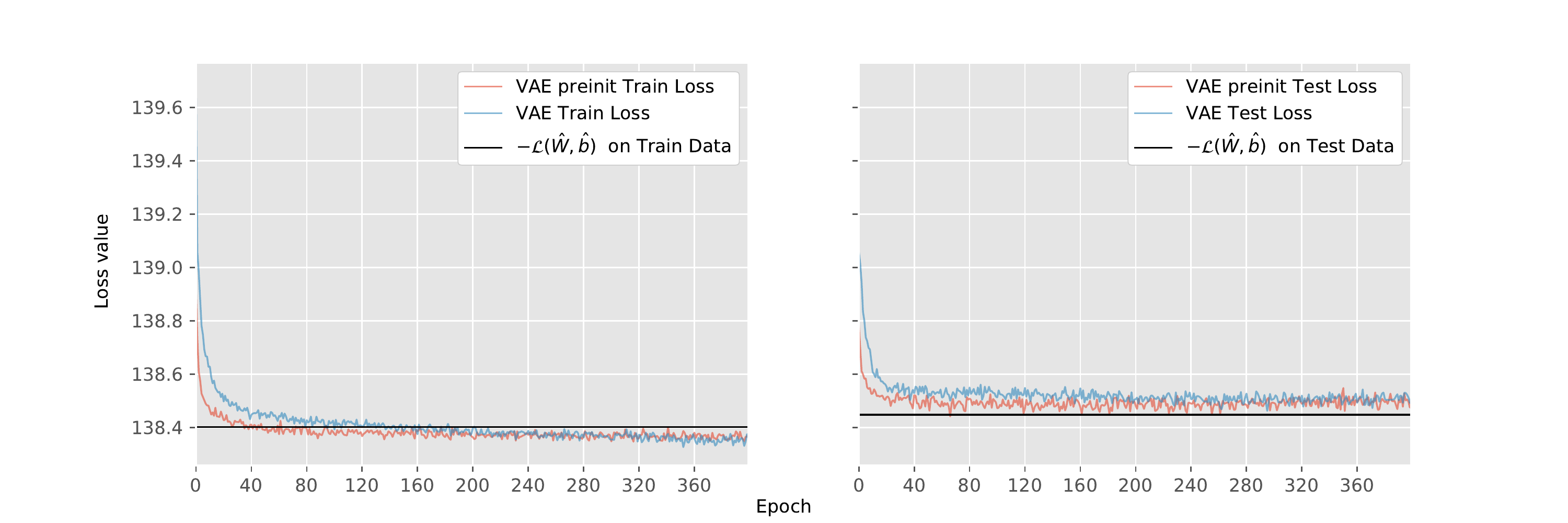}}
		\caption{The figure displays the behaviour of the loss of VAEs as constructed in \ref{subsec:VAE architecture} over 400 epochs of training. The data was generated as in \ref{subsec:Data generation} with $N=10000$ and $d=200$. On the left we see the performance on the training set and on the right on the test set. The derived upper bound (for the negative of the average ELBO) is in both cases displayed as a black line.}
		\label{fig:vae_vae_preinit_paper}
	\end{center}
	\vskip -0.2in
\end{figure*}

Given the possible combinations of $d\in\{200;400;1000\}$ and $N \in \{100;5.000;10000\}$, we have nine different simulations. We generate data once and for each of the nine simulation settings we train the two VAEs with $100$ different (randomly) initialized starting points. The ``VAE preinit''  layers affected by our initialization keep their values.  For this net we get different (randomly) initialized starting points for the remaining layers.

Each simulation (200 net trainings) was executed on a Dual Intel Xeon Gold 6132 ("Skylake") @ 2.6 GHz with 28 CPU cores. The longest setup ($N=10000$ and $d=4000$) took about two and a half days of computing time. We used and modified the implementation of a VAE provided by \cite{keras}.\footnote{Our code together with a readme file for execution can be found in the provided file ``vae\_analysis\_binary.zip'' in the supplementary materials.}

Table \ref{tab:minmax test Deep} and table \ref{tab:minmax train Deep} display information on loss values at the end of $400$ epochs of training. We calculate the deviation from the bound value displayed in the same row and provide the interval of minimal and maximal deviation, displayed as ``[Min\%,Max\%]''.\footnote{To recover the real minimal or maximal value calculate $-\hELBO \cdot (1 + \textnormal{deviation in \%})$: e.g. $134.52 \cdot (1 - 16.02 \%) \approx 112.97$} The ``[Min\%,Max\%]'' intervals let us highlight the main aspects later on better than a ``mean $\pm$ std''-format. The mean and standard deviation values for the deep leaning architectures can be found in the supplementary material. We also present the simulation results (``[Min\%,Max\%]'' and ``mean $\pm$ std'') for the ``canonical'' architectures there.

In table \ref{tab:minmax test Deep} and table \ref{tab:minmax train Deep}, the nets perform better with smaller values. A negative percentage means that the final loss is smaller than the bound. As stated earlier, the deep architecture we look at here is actually not comparable to the bound. Hence, the loss values can be greater than the bound. 

Looking at the tables \ref{tab:minmax test Deep} and \ref{tab:minmax train Deep}, we draw the following four conclusions:
\begin{enumerate}
	\item Comparing the test cases, we favour ``VAE preinit'' in all but one setup $d=1000$ and $N=10000$.
	\item For the simulation setups with $N=100$ both nets are over-fitting. For all cases the ``VAE preinit'' is less prone to this behaviour, as can explicitly be seen from the performance on the test data in table \ref{tab:minmax test Deep}.
	\item For the training data, the bound is often reached by both nets. For the test data though, the resulting loss is likely to lie above the bound.
	\item Overall, the intervals are getting tighter for increasing $N$: Not surprisingly more data results in better fits. The bound is more robust: The values for different $N$ are comparable.
\end{enumerate}

\begin{table}[t]
	\caption{The table shows performances on the test data for deep learning architectures. For each net, we look at the information of 100 loss values after 400 epochs of training. The values displayed are the minimal and maximal relative deviations from the bound provided in the same row. They are displayed as ``[Min\%,Max\%]''.}
	\label{tab:minmax test Deep}
	\vskip 0.15in
	\begin{center}
		\begin{small}
			\begin{sc}
				\setlength{\extrarowheight}{10pt}
				\scalebox{0.85}{
					\begin{tabular}{cccc}
						\toprule
						Sim. & $-\hELBO$& \multicolumn{1}{l}{VAE preinit} & \multicolumn{1}{l}{VAE} \\ 
						\midrule
						\makecell{d=200;\\ N=100 } & \textbf{139.34} & [5.66\%;12.66\%] & [9.02\%;22.08\%] \\ 
						\makecell{d=200;\\ N=5000 } & \textbf{138.48} & [0.01\%;0.11\%] & [0.10\%;0.20\%] \\ 
						\makecell{d=200;\\ N=10000} & \textbf{138.45} & [-0.01\%;0.05\%] & [0.02\%;0.07\%] \\ 
						\midrule
						\makecell{d=400;\\ N=100 } & \textbf{279.08} & [5.69\%;12.68\%] & [7.97\%;17.19\%] \\ 
						\makecell{d=400;\\ N=5000 } & \textbf{277.22} & [0.04\%;0.09\%] & [0.24\%;0.31\%] \\ 
						\makecell{d=400;\\ N=10000} & \textbf{277.11} & [0.02\%;0.05\%] & [0.02\%;0.11\%] \\ 
						\midrule
						\makecell{d=1000;\\ N=100 } & \textbf{694.80} & [5.55\%;11.81\%] & [6.92\%;15.33\%] \\ 
						\makecell{d=1000;\\ N=5000 } & \textbf{693.25} & [0.08\%;0.13\%] & [0.02\%;0.35\%] \\ 
						\makecell{d=1000;\\ N=10000} & \textbf{693.12} & [0.02\%;0.04\%] & [0.00\%;0.02\%] \\ 
						\bottomrule
					\end{tabular}
				}
			\end{sc}
		\end{small}
	\end{center}
	\vskip -0.1in
\end{table}

\begin{table}[t]
	\caption{The table shows performances on the training data for deep learning architectures. For a description see the caption of table \ref{tab:minmax test Deep}.}
	\label{tab:minmax train Deep}
	\vskip 0.15in
	\begin{center}
		\begin{small}
			\begin{sc}
				\setlength{\extrarowheight}{10pt}
				\scalebox{0.85}{
					\begin{tabular}{cccc}
						\toprule
						Sim. & $-\hELBO$& \multicolumn{1}{l}{VAE preinit} & \multicolumn{1}{l}{VAE} \\ 
						\midrule
						\makecell{d=200;\\ N=100} & \textbf{134.52} & [-16.02\%;-11.58\%] & [-25.76\%;-21.81\%] \\ 
						\makecell{d=200;\\ N=5000 } & \textbf{138.35} & [-0.09\%;-0.03\%] & [-0.13\%;-0.06\%] \\ 
						\makecell{d=200;\\ N=10000} & \textbf{138.39} & [-0.04\%;-0.01\%] & [-0.06\%;-0.02\%] \\ 
						\midrule
						\makecell{d=400;\\ N=100 } & \textbf{269.59} & [-22.43\%;-17.28\%] & [-29.72\%;-24.42\%] \\ 
						\makecell{d=400;\\ N=5000 } & \textbf{276.91} & [-0.25\%;-0.14\%] & [-0.41\%;-0.30\%] \\ 
						\makecell{d=400;\\ N=10000} & \textbf{276.99} & [-0.06\%;-0.03\%] & [-0.10\%;0.01\%] \\ 
						\midrule
						\makecell{d=1000;\\ N=100 } & \textbf{674.44} & [-27.56\%;-22.05\%] & [-32.76\%;-27.03\%] \\ 
						\makecell{d=1000;\\ N=5000 } & \textbf{692.54} & [-0.59\%;-0.44\%] & [-0.55\%;0.00\%] \\ 
						\makecell{d=1000;\\ N=10000} & \textbf{692.75} & [-0.19\%;-0.04\%] & [0.02\%;0.05\%] \\ 
						\bottomrule
					\end{tabular}
				}
			\end{sc}
		\end{small}
	\end{center}
	\vskip -0.1in
\end{table}

\subsubsection{Loss development for deep architectures}

Figure \ref{fig:vae_vae_preinit_paper} displays loss values of the two different VAEs for test and training data. Furthermore, the bound $-\hELBO(\hat{W},\hat{b})$ was calculated for these two sets and added to the figure. As we are looking at the losses (the negative average ELBOs) of neural nets, smaller is better. 
We note that the ``VAE preinit'' has a faster convergence for both training and test set. After $400$ epochs, the different net architectures produce comparable losses. We conclude that fixing the ``$\log\bsigma_z^2$''-layer in ``VAE preinit'' produces a comparable final loss value. 
For the training set, both net architectures admit the bound $-\hELBO(\hat{W},\hat{b})$ after enough steps of training. Even though our theory from section \ref{sec:A lower bound for the ELBO} produces this bound for less parametrized net-structures than we look at here.

\section{Conclusion and possible extensions}
\label{sec:Conclusion and possible extensions}

The derived theoretical bound for VAE is easy to calculate and we show that even the performance of deep VAE-architectures (six layers) can be assessed with it. Therefore, we propose to use our bound to monitor the training of Bernoulli VAE. \\
Using our proposed initialization and the modification of the Bernoulli VAE architecture, a faster convergence rate is visible. Furthermore, we get more favourable results for the final loss values in all but one of our simulations. As a result, we motivate practitioners with real world binary data to apply a Bernoulli VAE with this initialization and architecture to their data. \\
The decision support for the optimal latent dimension $\kappa$ reduces the set of possible values when using a PCA.


%
%


\bibliography{BernoulliVAE.bib}
\bibliographystyle{apalike} 

\clearpage

\appendix
\section{Supplementary Material}
\subsection{Proof of proposition \ref{wohoo-i-got-a-proposition}}

\begin{proof}
	To proof proposition \ref{wohoo-i-got-a-proposition}, we will change the perspective. Instead of maximizing the ELBO, we want to minimize the negative ELBO given by
	\begin{align}\nonumber
	-\ELBO(\bphi,\btheta) :=& \dfrac{1}{N} \sum_{i=1}^N \KL\left(q_{\bphi}(Z|\ix)|| P(Z)\right) \\
	&- \E_{Z \sim q_{\bphi}\left(\cdot|\ix\right)}\left[\log P_{\btheta}(\ix|Z)\right] .\label{eq:bin-ELBO}
	\end{align}
	
	Looking at \eqref{eq:bin-ELBO}, we see two terms.
	For the KL-Divergence we have that
	\begin{align}\label{KL_equiv}
	&2 \cdot \KL\left(q_{\bphi}(Z|\ix)|| P(Z)\right) \nonumber\\
	&= \tr[\bSigmazi] - \log |\bSigmazi| + ||\bmuzi||_2^2 - \kappa
	\end{align}
	and for the second term (with $q_{\bphi}$ as abbreviation for $q_{\bphi}(\cdot|\ix)$) we derive that
	\begin{align}
	&- \E_{q_{\bphi}}\left[\log P_{\btheta}(\ix|Z)\right] \nonumber\\&= - \E_{q_{\bphi}}\left[\log  \prod_{j=1}^{d}\bigg\{ p_{\btheta,j}(z)^{\ix_j} (1-p_{\btheta,j}(z))^{(1- \ix_j)} \bigg\}\right]& \nonumber\\
	&= \E_{q_{\bphi}}\left[-{\ix}^T (WZ + b)\right] \nonumber\\
	&- \sum_{j=1}^{d} \E_{q_{\bphi}}\left[ \log \left(1-\sigma\left(\left[WZ + b\right]_j\right)\right)\right]. &\label{problemstelle von bin-ELBO}
	\end{align}
	For the first term we have 
	\begin{equation}\label{the-first-exp}
	\E_{q_{\bphi}}[-{\ix}^T (WZ + b)]=-{\ix}^T (W\bmuzi + b).
	\end{equation}
	If we do a Taylor series expansion in the point $z_0 = 0$ for the function $g(z):=\log \left(1-\sigma(z)\right)$ we get
	\[
	g(z) = -\log(2) - \dfrac{z}{2} - \dfrac{z^2}{8} + R_2(g,0;z).
	\]
	\begin{figure}
		\centering
		\includegraphics[width=0.5\linewidth]{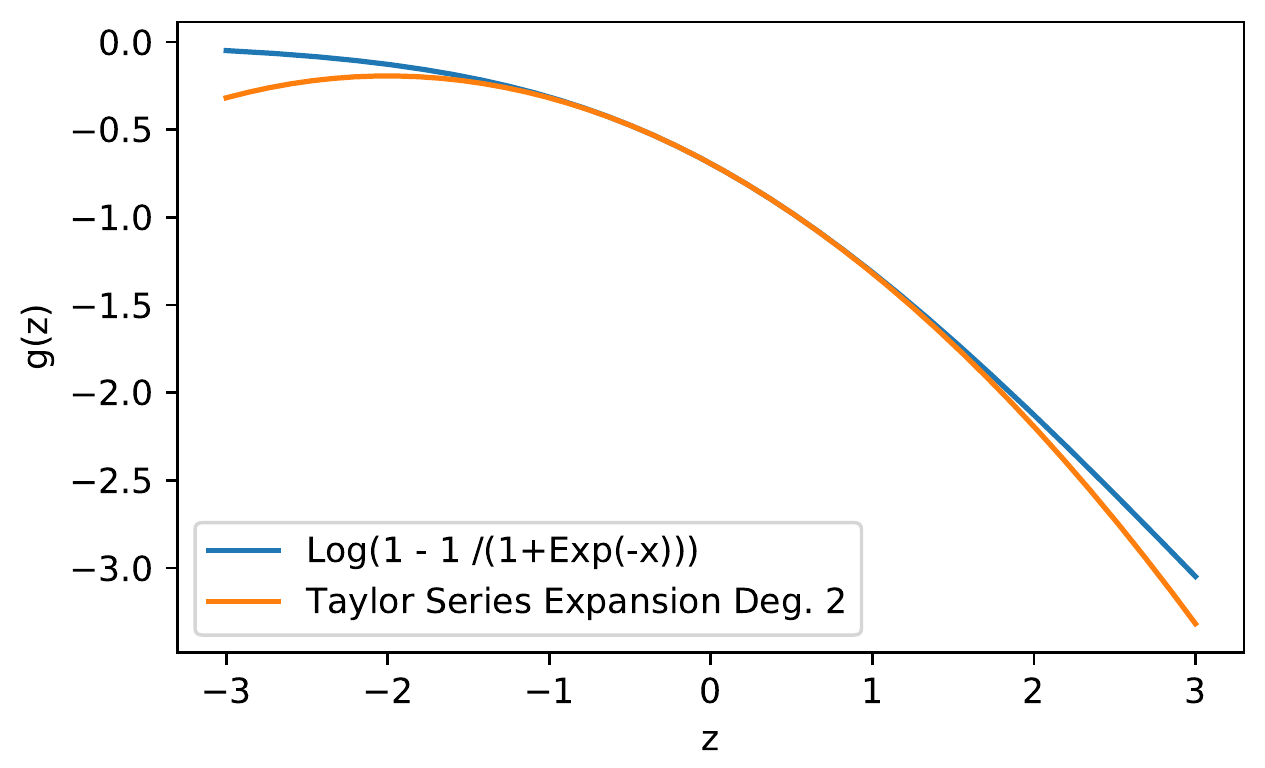}
		\caption{For the range -3 to 3 the Taylor series expansion of the function $\log(1- \sigma(z))$, where $\sigma(\cdot)$ denotes the sigmoid function. The desired function is approximated from below by the degree 2 expansion. }\label{fig:Taylor_Series_Expansion}
	\end{figure}
	This approximation is shown in figure \ref{fig:Taylor_Series_Expansion}. A degree 2 expansion seems reasonable and we see that it is approaching the original function from below. In fact, we have that for the remainder $R_2(g,0;z)$ it holds
	\[
	R_2(g,0;z) \geq 0 \quad \forall z \in \R.
	\]
	An detailed argument is given section \ref{subsec:Analysing the Taylor Remainder}. 
	
	Leaving the remainder aside, we get for the expectation in \eqref{problemstelle von bin-ELBO} that
	
	\begin{align*}
	&\E_{q_{\bphi}}\bigg(\log \left( 1- \sigma\left(\left[WZ + b\right]_j\right)\right)\bigg) \geq\\ 
	&-\log(2) - \dfrac{\E_{q_{\bphi}}([WZ + b]_j)}{2} - \dfrac{\E_{q_{\bphi}}([WZ + b]_j^2)}{8}. 
	\end{align*}
	We can write $Z=\bSzi Y + \bmuzi$, with $Y \sim \Nor(\bzero, \bI)$. Therefore, if $\Wj$ denotes the $j$-th row of $W$, for
	\[
	[WZ + b]_j = \Wj\bSzi Y + \Wj\bmuzi + b_j
	\] 
	it follows that 
	\begin{equation}\label{fist_moment}
	\E_{q_{\bphi}}([WZ + b]_j) = \Wj\bmuzi + b_j
	\end{equation}
	and
	\begin{equation}\label{second}
	\E_{q_{\bphi}}([WZ + b]_j^2) =  \Wj \bSigmazi \Wj^T + (\Wj\bmuzi + b_j)^2.
	\end{equation}
	For the sum in \eqref{problemstelle von bin-ELBO}, we get
	\begin{align*}
	&\sum_{j=1}^{d} \E_{q_{\bphi}}[ \log \sigma([WZ + b]_j)] \\
	&\geq - d\log(2) - \dfrac{1}{2} \Ind^TW\bmuzi - \dfrac{1}{2} \Ind^Tb\\
	&- \dfrac{1}{8} \tr(W\bSigmazi W^T) - \dfrac{1}{8} ||W\bmuzi||_2^2 \\
	&- \dfrac{1}{4} b^T W\bmuzi - \dfrac{1}{8}  ||b||_2^2,
	\end{align*}
	with $\Ind= (1,\ldots,1)^T \in \R^d$.
	We always approximate the sum from below and hence the negative ELBO from above.
	Putting these results together with \eqref{KL_equiv} and \eqref{the-first-exp} for our target function in \eqref{eq:bin-ELBO}, it follows that it is smaller than
	\begin{align*}
	-\hELBO(\bphi,W,b):=
	&\dfrac{1}{N} \sum_{i=1}^N \Bigg[ \dfrac{\tr[\bSigmazi]}{2} - \dfrac{\log |\bSigmazi|}{2} + \dfrac{||\bmuzi||_2^2}{2}\\
	&- \dfrac{\kappa}{2}- {\ix}^T (W\bmuzi + b)\\
	&+ d\log(2) + \dfrac{1}{2} \Ind^TW\bmuzi + \dfrac{1}{2} \Ind^Tb\\
	&+ \dfrac{1}{8} \tr(W\bSigmazi W^T) + \dfrac{1}{8} ||W\bmuzi||_2^2 \\
	&+ \dfrac{1}{4} b^T W\bmuzi + \dfrac{1}{8}  ||b||_2^2 \Bigg].
	\end{align*}
	
	All potential minima w.r.t. $\bSigmazi$ have to conform to 
	\begin{equation}\label{opt-bSigmaz}
	\bhSigma_z := \bhSigmazi = (I_{\kappa} + \dfrac{1}{4} W^TW)^{-1},
	\end{equation}
	independent of $\ix$. To see that these are also minima and not maxima, consider that it holds (see \ref{subsec:The maximum Entropy completion problem}) that
	\[
	\kappa + \log |A A^T|= \min_{\Gamma \succ 0} \tr(AA^T \Gamma^{-1}) + \log |\Gamma|,
	\]
	where we set $\Gamma^{-1} = \bSigma_z$ and $AA^T = I_{\kappa} + \dfrac{1}{4} W^TW$. 
	
	Given $\bhSigma_z$ the candidates for the minimal target function are writeable as 
	\begin{align*}
	-\hELBO(\bphi \setminus \{\bSigma_z\},W,b) =  &\dfrac{1}{N}\sum_{i=1}^N \Bigg[ \left(-{\ix} + \dfrac{1}{2} \Ind+ \dfrac{1}{4} b\right)^T W\bmuzi + \dfrac{||\bmuzi||_2^2}{2} \\
	&+ \dfrac{1}{8} ||W\bmuzi||_2^2
	+ (-{\ix}+ \dfrac{1}{2} \Ind)^Tb + \dfrac{1}{8}  ||b||_2^2 \\ &+\dfrac{1}{2}\log|I_{\kappa} + \dfrac{1}{4} W^TW| + d\log(2) \Bigg].
	\end{align*}
	
	For $\bmuzi$, we achieve as minimal points
	\begin{align}
	\bhmuzi &= \bhSigma_z W^T\Big(\ix - \dfrac{1}{2} \Ind- \dfrac{1}{4} b\Big) \nonumber \\
	&= \bhSigma_z W^T\Big(\iy - \dfrac{1}{4} b\Big)\label{opt-bmuzi}
	\end{align}
	with $\iy := \Big(\ix - \dfrac{1}{2} \Ind\Big) \in \bigg\{-\dfrac{1}{2},\dfrac{1}{2}\bigg\}^d$. These are minima since the second derivative is $\bhSigma^{-1}$, which is positive definite. Given the optimal $\bmuzi$ and $\bSigmazi$ our target function is only dependent on the parameters $W$ and $b$. Hence
	\begin{align}
	-\hELBO(W,b) =  &\dfrac{1}{N}\sum_{i=1}^N \Bigg[ \Big(\iy - \dfrac{1}{4} b\Big)^T E \Big(\iy - \dfrac{1}{4} b\Big) \label{SVD-prob}
	- {\iy}^Tb \Bigg] \\
	&+ \dfrac{1}{8}  ||b||_2^2 +\dfrac{1}{2}\log|I_{\kappa} + \dfrac{1}{4} W^TW| + d\log(2), \nonumber
	\end{align}
	where $E:=W \Big(\dfrac{1}{2}\bhSigma_z^2 - \bhSigma_z +\dfrac{1}{8} \bhSigma_z W^TW \bhSigma_z\Big) W^T$ and $\bhSigma_z$ is fixed. 
	Consider the Singular Value Decomposition of $W = U \widetilde{D} V^T$ with $U\in \R^{d\times d}$ and $V\in \R^{\kappa\times \kappa}$ are unitary matrices and 
	\[
	\widetilde{D} = \begin{bmatrix}
	d_1 & \ldots & 0 \\
	\vdots & \ddots & \vdots \\
	0 & \ldots & d_{\kappa} \\
	\vdots & \ddots & \vdots \\
	0 & \ldots & 0 \\
	\end{bmatrix}\in \R^{d\times \kappa}.
	\]
	
	We have 
	\[
	\widetilde{D}^T \widetilde{D} = \begin{bmatrix}
	d_1^2 & \ldots & 0 \\
	\vdots & \ddots & \vdots \\
	0 & \ldots & d_{\kappa}^2 \\
	\end{bmatrix}
	\]
	and can therefore write 
	\begin{align*}
	\bhSigma_z =&\Big(V (I_{\kappa} + \dfrac{1}{4} \widetilde{D}^T \widetilde{D})  V^T\Big)^{-1} \\
	&= V \widehat{D} V^T,
	\end{align*}
	with $\widehat{D} := \diag\left(\dfrac{4}{4 + d_1^2} ,\ldots, \dfrac{4}{4 + d_{\kappa}^2}\right)$. For the matrix expression in the middle of the term in \eqref{SVD-prob} it follows that 
	\begin{align}
	&W \Big(\dfrac{1}{2}\bhSigma_z^2 - \bhSigma_z +\dfrac{1}{8} \bhSigma_z W^TW \bhSigma_z\Big) W^T \nonumber \\
	&=  W \Big(V \Big(\dfrac{1}{2}\widehat{D}^2 - \widehat{D} +\dfrac{1}{8} \widehat{D} \widetilde{D}^T \widetilde{D} \widehat{D} \Big)V^T\Big) W^T  \nonumber\\
	&= W \Big(V \breve{D} V^T\Big) W^T, \label{SVD-prob-part1} 
	\end{align}
	where we denote $\breve{D} = \diag\left(\dfrac{-2}{4 + d_1^2} ,\ldots, \dfrac{-2}{4 + d_{\kappa}^2}\right)$. The justification of the last equation becomes apparent, when we consider one respective diagonal element $d_{\cdot}$ of the diagonal matrices in the equation. We have
	\begin{align*}
	&\dfrac{1}{2}\dfrac{16}{(4 + d_{\cdot}^2)^2} - \dfrac{4}{4 + d_{\cdot}^2} + \dfrac{1}{8}\dfrac{16 d_{\cdot}^2}{(4 + d_{\cdot}^2)^2} \\  
	&= \dfrac{8}{(4 + d_{\cdot}^2)^2} - \dfrac{16 + 4 d_{\cdot}^2}{(4 + d_{\cdot}^2)^2} + \dfrac{2 d_{\cdot}^2}{(4 + d_{\cdot}^2)^2} \\
	&=\dfrac{-8 - 2 d_{\cdot}^2}{(4 + d_{\cdot}^2)^2} \\
	&=\dfrac{-2}{(4 + d_{\cdot}^2)}.
	\end{align*}
	We can further rewrite \eqref{SVD-prob-part1} as
	\begin{align*}
	W \Big(V \breve{D} V^T\Big) W^T &= U \widetilde{D} \breve{D} \widetilde{D}^T U^T \\
	&=: -U D U^T ,
	\end{align*}
	where we have introduced 
	\[
	D:=\begin{bmatrix}
	\diag\left(\dfrac{2d_1^2}{4 + d_1^2}, \ldots,\dfrac{2d_{\kappa}^2}{4 + d_{\kappa}^2} \right) & \bO \\
	\bO & \bO
	\end{bmatrix} \in \R^{d \times d}.
	\]
	So for our target function in \eqref{SVD-prob}, we get
	\begin{align*}
	-\hELBO(W,b) =  &\dfrac{1}{N}\sum_{i=1}^N \Bigg[ -\Big(\iy - \dfrac{1}{4} b\Big)^TU D U^T\Big(\iy - \dfrac{1}{4} b\Big)
	- {\iy}^Tb \Bigg] \\
	& + \dfrac{1}{8}  ||b||_2^2 +\dfrac{1}{2}\log|I_{\kappa} + \dfrac{1}{4} W^TW|+ d\log(2). 
	\end{align*}
	
	By adding and subtracting the constant term $\dfrac{1}{N}\sum_{i=1}^N 2 ||\iy||_2^2 = \dfrac{d}{2}$, concluding 
	\[
	2 ||\iy||_2^2 - \dfrac{4}{4}{\iy}^Tb + \dfrac{2}{8 \cdot 2} ||b||_2^2 = 2 \cdot (\iy - \dfrac{1}{4} b)^T(\iy - \dfrac{1}{4} b)
	\]
	and together with
	\[
	\dfrac{1}{2}\log|I_{\kappa} + \dfrac{1}{4} W^TW| + d \log(2)  = \dfrac{1}{2}\log|4 I_{d} + WW^T|,
	\]
	we get 
	\begin{align*}
	-\hELBO(W,b) =  &\dfrac{1}{N}\sum_{i=1}^N \Bigg[ \Big(\iy - \dfrac{1}{4} b\Big)^TU ( 2I_{d} - D) U^T\Big(\iy - \dfrac{1}{4} b\Big)\Bigg] \\
	&+ \dfrac{1}{2}\log|4 I_{d} + WW^T| - \dfrac{d}{2}.
	\end{align*}
	We can further rewrite 
	\[
	U(2I_{d} - D)U^T = \dfrac{16}{2} (4I_{d} + WW^T)^{-1}
	\] and get
	\begin{align*}
	-\hELBO(W,b) =  &\dfrac{1}{N}\sum_{i=1}^N \Bigg[ \dfrac{1}{2} \Big(4 \iy - b\Big)^T (4I_{d} + WW^T)^{-1} \Big( 4 \iy - b\Big) \Bigg]\\
	& + \dfrac{1}{2}\log|4 I_{d} + WW^T| - \dfrac{d}{2}.
	\end{align*}
\end{proof}

\subsection{Analysing the Taylor Remainder}\label{subsec:Analysing the Taylor Remainder}

Let's have a look at the target term that we introduced when using a Taylor series expansion. The term we were considering is given by
\begin{equation*}
\sum_{j=1}^{d}\E_{q_{\bphi}}\left[\log\left(1- \sigma\left(\left[WZ+b\right]_j\right)\right)\right].
\end{equation*}
The Taylor series expansion in the point $z_0 = 0$ for the function $g(z):=\log \left(1-\sigma(z)\right)$ is
\[
\log \sigma(z) = -\log(2) - \dfrac{z}{2} - \dfrac{z^2}{8} + R_2(g,0;z),
\]
where $R_2(g,0;z)$ is the remainder of the expansion. It can be explicitly calculated as Lagrange Remainder

\begin{equation}\label{eq:LagRemainder}
R_2(g,0;z) = \dfrac{g^{(3)}(\xi)}{6} z^3
\end{equation}

with $g^{(3)}(\xi) = \dfrac{e^z (e^z -1)}{(1+e^z)^3}$ denoting the third derivative. $\xi$ is a value between the expansion point $0$ and $z$.
\begin{figure}
	\centering
	\includegraphics[width=0.5\linewidth]{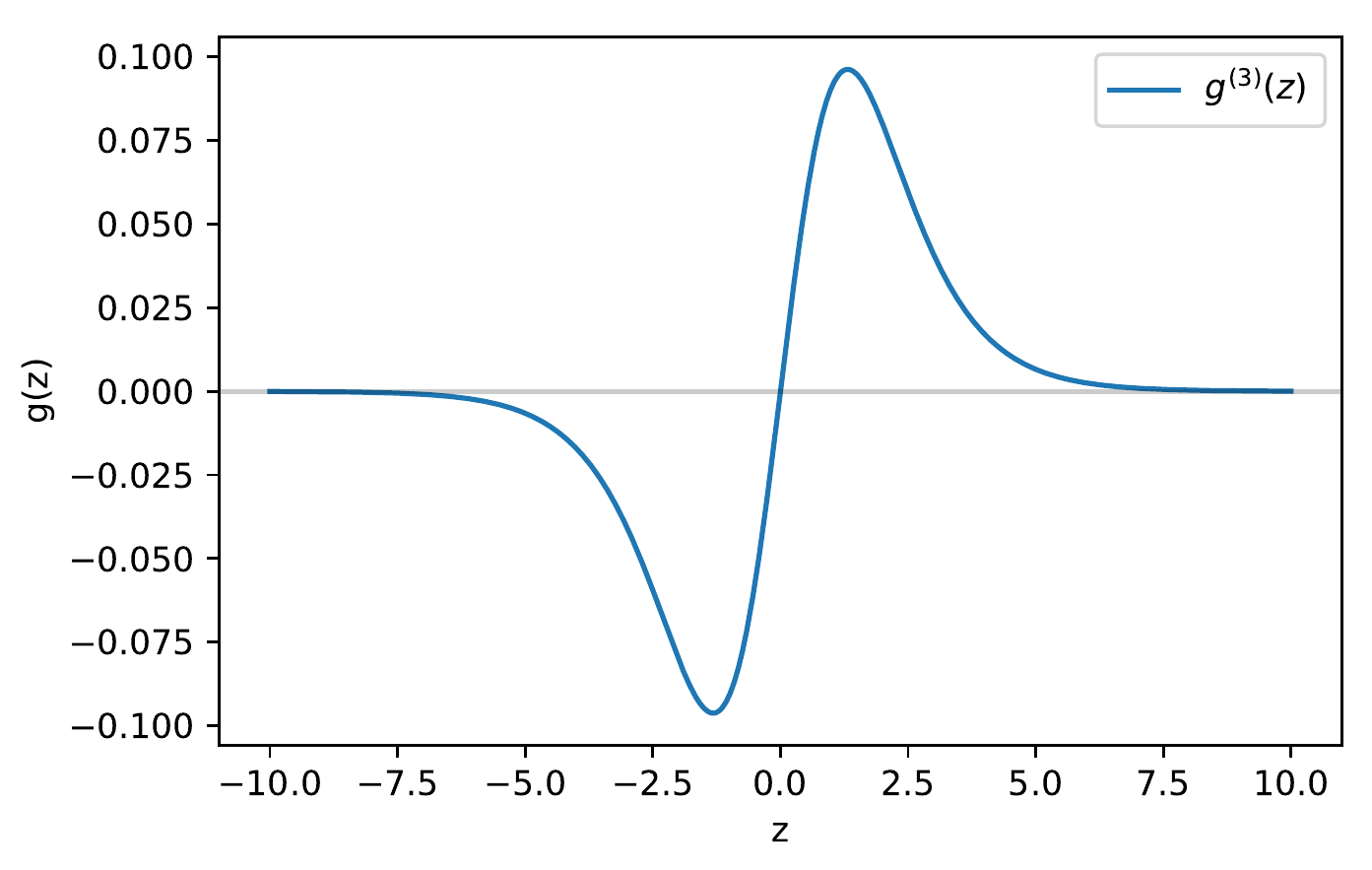}
	\caption{The third derivative of the function $g(z):=\log \sigma(z)$. It is point symmetric: For negative values it is negative and for positive values it is positive. The maximum/minimum is achieved at $\log(2\pm \sqrt{3})$ and given by $z = \pm \dfrac{1}{6\sqrt{3}}$.}\label{fig:TayRemThirdDerivative}
\end{figure}
This derivative is plotted in figure \ref{fig:TayRemThirdDerivative}. It is point symmetric: For negative values it is negative and for positive values it is positive. Hence, the error that we make is always positive!

\subsection{The maximum Entropy completion problem}\label{subsec:The maximum Entropy completion problem}
Let $B,\Gamma \in \R^{\kappa \times \kappa}$ be symmetric positive definite matrices. 
Then it holds
\[
B = \arg\min_{\Gamma \succ 0}  \tr(B \Gamma^{-1}) + \log |\Gamma|.
\]
and hence
\[
\kappa  + \log |B| = \min_{\Gamma \succ 0}  \tr(B \Gamma^{-1}) + \log |\Gamma|.
\] 
\begin{proof}
	Define the two distributions $\Nor_0(\bmu,B)$ and  $\Nor_1(\bmu,\Gamma)$. We have
	\begin{align}\label{max_ent}
	&2 \cdot \KL\left(\Nor_0(\bmu,B)|| \Nor_1(\bmu,\Gamma)\right) = \tr(B \Gamma^{-1}) + \log |\Gamma| - \kappa -  \log |B| 
	\end{align}
	Now, consider that for the Kullback-Leibler-Divergence with probability distributions $P$ and $Q$ it holds:
	\begin{itemize}
		\item $\KL\left(  P || Q \right) \geq 0$ for all inputs.
		\item $\KL\left( P || Q\right) = 0$ if and only if $P=Q$ almost everywhere.
	\end{itemize}
	Hence, we conclude $B=\Gamma$ in the minimum.
\end{proof}

\subsection{Simulation results for ``canonical'' architectures}

Here, the test and training data results are provided for following cases: 
\begin{itemize}
	\item The ``canonical'' architecture with ``[Min\%,Max\%]'' format in table \ref{tab: minmax test canon} and \ref{tab: minmax train canon}.
	\item The ``canonical'' architecture with ``mean $\pm$ std'' format in table \ref{tab: meanstd test canon} and \ref{tab: meanstd train canon}.
	\item The deep architecture with ``mean $\pm$ std'' format in table \ref{tab: meanstd test Deep} and \ref{tab: meanstd train Deep}.
\end{itemize}
For a description of the ``[Min\%,Max\%]''-format see section \ref{subsubsec:Simulation results for the deep architectures}. The ``mean $\pm$ std''-format presents the mean relative deviation (in \%) and the std of this value.

The value of $-\hELBO$ in the tables of ``canonical''- architecture is different to the deep architecture due to a different seed during simulation.

\begin{table}[t]
	\caption{The table shows performances on the test data for ``canonical'' architectures in the ``[Min\%,Max\%]''-format. For a description, see table \ref{tab:minmax test Deep} and section \ref{subsubsec:Simulation results for the deep architectures}.}
	\label{tab: minmax test canon}
	\vskip 0.15in
	\begin{center}
		\begin{small}
			\begin{sc}
				\setlength{\extrarowheight}{10pt}
				\scalebox{0.85}{
					\begin{tabular}{cccc}
						\toprule
						Sim. & $-\hELBO$& \multicolumn{1}{l}{VAE preinit} & \multicolumn{1}{l}{VAE} \\ 
						\midrule
						\makecell{d=200;\\ N=100} & \textbf{138.94} & [1.71\%;2.47\%] & [-0.20\%;-0.05\%] \\ 
						\makecell{d=200;\\ N=5000 } & \textbf{138.49} & [-0.03\%;0.05\%] & [-0.01\%;0.07\%] \\ 
						\makecell{d=200;\\ N=10000} & \textbf{138.46} & [-0.02\%;0.04\%] & [-0.01\%;0.06\%] \\ 
						\midrule
						\makecell{d=400;\\ N=100 } & \textbf{278.53} & [1.22\%;1.61\%] & [-0.49\%;-0.38\%] \\ 
						\makecell{d=400;\\ N=5000 } & \textbf{277.21} & [0.00\%;0.04\%] & [0.00\%;0.09\%] \\ 
						\makecell{d=400;\\ N=10000} & \textbf{277.13} & [0.00\%;0.02\%] & [0.01\%;0.03\%] \\ 
						\midrule
						\makecell{d=1000;\\ N=100 } & \textbf{695.35} & [1.11\%;1.29\%] & [-0.31\%;-0.24\%] \\ 
						\makecell{d=1000;\\ N=5000 } & \textbf{693.29} & [0.02\%;0.03\%] & [0.01\%;0.04\%] \\ 
						\makecell{d=1000;\\ N=10000} & \textbf{693.13} & [0.00\%;0.02\%] & [0.00\%;0.03\%] \\  
						\bottomrule
					\end{tabular}
				}
			\end{sc}
		\end{small}
	\end{center}
	\vskip -0.1in
\end{table}

\begin{table}[t]
	\caption{The table shows performances on the training data for ``canonical'' architectures in the ``[Min\%,Max\%]''-format. For a description, see table \ref{tab:minmax test Deep} and section \ref{subsubsec:Simulation results for the deep architectures}.}
	\label{tab: minmax train canon}
	\vskip 0.15in
	\begin{center}
		\begin{small}
			\begin{sc}
				\setlength{\extrarowheight}{10pt}
				\scalebox{0.85}{
					\begin{tabular}{cccc}
						\toprule
						Sim. & $-\hELBO$& \multicolumn{1}{l}{VAE preinit} & \multicolumn{1}{l}{VAE} \\ 
						\midrule
						\makecell{d=200;\\ N=100} & \textbf{134.83} & [-1.52\%;-0.95\%] & [2.47\%;2.67\%] \\ 
						\makecell{d=200;\\ N=5000 } & \textbf{138.35} & [-0.06\%;0.00\%] & [-0.03\%;0.05\%] \\ 
						\makecell{d=200;\\ N=10000} & \textbf{138.40} & [-0.04\%;0.00\%] & [-0.03\%;0.03\%] \\ 
						\midrule
						\makecell{d=400;\\ N=100 } & \textbf{268.46} & [-1.39\%;-1.16\%] & [2.80\%;2.98\%] \\ 
						\makecell{d=400;\\ N=5000 } & \textbf{276.90} & [-0.04\%;-0.01\%] & [-0.02\%;0.05\%] \\ 
						\makecell{d=400;\\ N=10000} & \textbf{276.99} & [-0.02\%;0.00\%] & [-0.02\%;0.01\%] \\ 
						\midrule
						\makecell{d=1000;\\ N=100 } & \textbf{674.50} & [-1.27\%;-1.17\%] & [2.10\%;2.25\%] \\ 
						\makecell{d=1000;\\ N=5000 } & \textbf{692.55} & [-0.03\%;-0.01\%] & [-0.01\%;0.01\%] \\ 
						\makecell{d=1000;\\ N=10000} & \textbf{692.76} & [-0.02\%;-0.01\%] & [-0.01\%;0.01\%] \\ 
						\bottomrule
					\end{tabular}
				}
			\end{sc}
		\end{small}
	\end{center}
	\vskip -0.1in
\end{table}

\begin{table}[t]
	\caption{The table shows performances on the test data for ``canonical'' architectures in the ``mean $\pm$ std''-format. For a description, see table \ref{tab:minmax test Deep} and section \ref{subsubsec:Simulation results for the deep architectures}.}
	\label{tab: meanstd test canon}
	\vskip 0.15in
	\begin{center}
		\begin{small}
			\begin{sc}
				\setlength{\extrarowheight}{10pt}
				\scalebox{0.85}{
					\begin{tabular}{cccc}
						\toprule
						Sim. & $-\hELBO$& \multicolumn{1}{l}{VAE preinit} & \multicolumn{1}{l}{VAE} \\ 
						\midrule
						\makecell{d=200;\\ N=100} & \textbf{138.94} & 2.02\% $\pm $0.15\% & -0.12\% $\pm $0.03\% \\ 
						\makecell{d=200;\\ N=5000 } & \textbf{138.49} & 0.01\% $\pm $0.02\% & 0.03\% $\pm $0.02\% \\ 
						\makecell{d=200;\\ N=10000} & \textbf{138.46} & 0.01\% $\pm $0.01\% & 0.02\% $\pm $0.01\% \\
						\midrule 
						\makecell{d=400;\\ N=100 } & \textbf{278.53} & 1.40\% $\pm $0.08\% & -0.43\% $\pm $0.02\% \\ 
						\makecell{d=400;\\ N=5000 } & \textbf{277.21} & 0.03\% $\pm $0.01\% & 0.02\% $\pm $0.01\% \\ 
						\makecell{d=400;\\ N=10000} & \textbf{277.13} & 0.01\% $\pm $0.01\% & 0.02\% $\pm $0.01\% \\
						\midrule 
						\makecell{d=1000;\\ N=100 } & \textbf{695.35} & 1.19\% $\pm $0.03\% & -0.28\% $\pm $0.02\% \\ 
						\makecell{d=1000;\\ N=5000 } & \textbf{693.29} & 0.02\% $\pm $0.00\% & 0.02\% $\pm $0.00\% \\ 
						\makecell{d=1000;\\ N=10000} & \textbf{693.13} & 0.01\% $\pm $0.00\% & 0.01\% $\pm $0.00\% \\ 
						\bottomrule
					\end{tabular}
				}
			\end{sc}
		\end{small}
	\end{center}
	\vskip -0.1in
\end{table}

\begin{table}[t]
	\caption{The table shows performances on the training data for ``canonical'' architectures in the ``mean $\pm$ std''-format. For a description, see table \ref{tab:minmax test Deep} and section \ref{subsubsec:Simulation results for the deep architectures}.}
	\label{tab: meanstd train canon}
	\vskip 0.15in
	\begin{center}
		\begin{small}
			\begin{sc}
				\setlength{\extrarowheight}{10pt}
				\scalebox{0.85}{
					\begin{tabular}{cccc}
						\toprule
						Sim. & $-\hELBO$& \multicolumn{1}{l}{VAE preinit} & \multicolumn{1}{l}{VAE} \\ 
						\midrule
						\makecell{d=200;\\ N=100} & \textbf{134.83} & -1.24\% $\pm $0.10\% & 2.55\% $\pm $0.04\% \\ 
						\makecell{d=200;\\ N=5000 } & \textbf{138.35} & -0.03\% $\pm $0.01\% & 0.02\% $\pm $0.01\% \\ 
						\makecell{d=200;\\ N=10000} & \textbf{138.40} & -0.02\% $\pm $0.01\% & 0.00\% $\pm $0.01\% \\ 
						\midrule
						\makecell{d=400;\\ N=100 } & \textbf{268.46} & -1.26\% $\pm $0.04\% & 2.92\% $\pm $0.04\% \\ 
						\makecell{d=400;\\ N=5000 } & \textbf{276.90} & -0.02\% $\pm $0.01\% & 0.00\% $\pm $0.01\% \\ 
						\makecell{d=400;\\ N=10000} & \textbf{276.99} & -0.01\% $\pm $0.00\% & -0.01\% $\pm $0.00\% \\ 
						\midrule
						\makecell{d=1000;\\ N=100 } & \textbf{674.50} & -1.23\% $\pm $0.02\% & 2.17\% $\pm $0.03\% \\ 
						\makecell{d=1000;\\ N=5000 } & \textbf{692.55} & -0.02\% $\pm $0.00\% & 0.00\% $\pm $0.01\% \\ 
						\makecell{d=1000;\\ N=10000} & \textbf{692.76} & -0.01\% $\pm $0.00\% & -0.01\% $\pm $0.00\% \\ 
						\bottomrule
					\end{tabular}
				}
			\end{sc}
		\end{small}
	\end{center}
	\vskip -0.1in
\end{table}

\begin{table}[t]
	\caption{The table shows performances on the test data for deep architectures in the ``mean $\pm$ std''-format. For a description, see table \ref{tab:minmax test Deep} and section \ref{subsubsec:Simulation results for the deep architectures}.}
	\label{tab: meanstd test Deep}
	\vskip 0.15in
	\begin{center}
		\begin{small}
			\begin{sc}
				\setlength{\extrarowheight}{10pt}
				\scalebox{0.85}{
					\begin{tabular}{cccc}
						\toprule
						Sim. & $-\hELBO$& \multicolumn{1}{l}{VAE preinit} & \multicolumn{1}{l}{VAE} \\ 
						\midrule
						\makecell{d=200;\\ N=100} & \textbf{139.34} & 8.60\% $\pm $1.37\% & 14.70\% $\pm $2.34\% \\ 
						\makecell{d=200;\\ N=5000 } & \textbf{138.48} & 0.06\% $\pm $0.02\% & 0.14\% $\pm $0.02\% \\ 
						\makecell{d=200;\\ N=10000} & \textbf{138.45} & 0.02\% $\pm $0.01\% & 0.05\% $\pm $0.01\% \\ 
						\midrule
						\makecell{d=400;\\ N=100 } & \textbf{279.08} & 8.64\% $\pm $1.46\% & 12.04\% $\pm $1.76\% \\ 
						\makecell{d=400;\\ N=5000 } & \textbf{277.22} & 0.07\% $\pm $0.01\% & 0.26\% $\pm $0.01\% \\ 
						\makecell{d=400;\\ N=10000} & \textbf{277.11} & 0.04\% $\pm $0.01\% & 0.09\% $\pm $0.02\% \\ 
						\midrule
						\makecell{d=1000;\\ N=100 } & \textbf{694.80} & 8.73\% $\pm $1.40\% & 10.58\% $\pm $1.82\% \\ 
						\makecell{d=1000;\\ N=5000 } & \textbf{693.25} & 0.09\% $\pm $0.01\% & 0.18\% $\pm $0.08\% \\ 
						\makecell{d=1000;\\ N=10000} & \textbf{693.12} & 0.03\% $\pm $0.00\% & 0.01\% $\pm $0.00\% \\ 
						\bottomrule
					\end{tabular}
				}
			\end{sc}
		\end{small}
	\end{center}
	\vskip -0.1in
\end{table}

\begin{table}[t]
	\caption{The table shows performances on the training data for deep architectures in the ``mean $\pm$ std''-format. For a description, see table \ref{tab:minmax test Deep} and section \ref{subsubsec:Simulation results for the deep architectures}.}
	\label{tab: meanstd train Deep}
	\vskip 0.15in
	\begin{center}
		\begin{small}
			\begin{sc}
				\setlength{\extrarowheight}{10pt}
				\scalebox{0.85}{
					\begin{tabular}{cccc}
						\toprule
						Sim. & $-\hELBO$& \multicolumn{1}{l}{VAE preinit} & \multicolumn{1}{l}{VAE} \\ 
						\midrule
						\makecell{d=200;\\ N=100} & \textbf{134.52} & -14.36\% $\pm $0.88\% & -23.45\% $\pm $0.77\% \\ 
						\makecell{d=200;\\ N=5000 } & \textbf{138.35} & -0.06\% $\pm $0.01\% & -0.10\% $\pm $0.01\% \\ 
						\makecell{d=200;\\ N=10000} & \textbf{138.39} & -0.03\% $\pm $0.01\% & -0.04\% $\pm $0.01\% \\ 
						\midrule
						\makecell{d=400;\\ N=100 } & \textbf{269.59} & -19.89\% $\pm $1.08\% & -27.44\% $\pm $0.96\% \\ 
						\makecell{d=400;\\ N=5000 } & \textbf{276.91} & -0.19\% $\pm $0.02\% & -0.36\% $\pm $0.02\% \\ 
						\makecell{d=400;\\ N=10000} & \textbf{276.99} & -0.05\% $\pm $0.01\% & -0.07\% $\pm $0.02\% \\ 
						\midrule
						\makecell{d=1000;\\ N=100 } & \textbf{674.44} & -24.71\% $\pm $1.08\% & -30.23\% $\pm $1.19\% \\ 
						\makecell{d=1000;\\ N=5000 } & \textbf{692.54} & -0.51\% $\pm $0.03\% & -0.24\% $\pm $0.14\% \\ 
						\makecell{d=1000;\\ N=10000} & \textbf{692.75} & -0.14\% $\pm $0.02\% & 0.04\% $\pm $0.01\% \\ 
						\bottomrule
					\end{tabular}
				}
			\end{sc}
		\end{small}
	\end{center}
	\vskip -0.1in
\end{table}

\subsection{Loss development for ``canonical'' net architectures}\label{subsec:Loss development for ``canonical'' net architectures}

\begin{figure*}[ht]
	\vskip 0.2in
	\begin{center}
		\centerline{\includegraphics[width=\columnwidth]{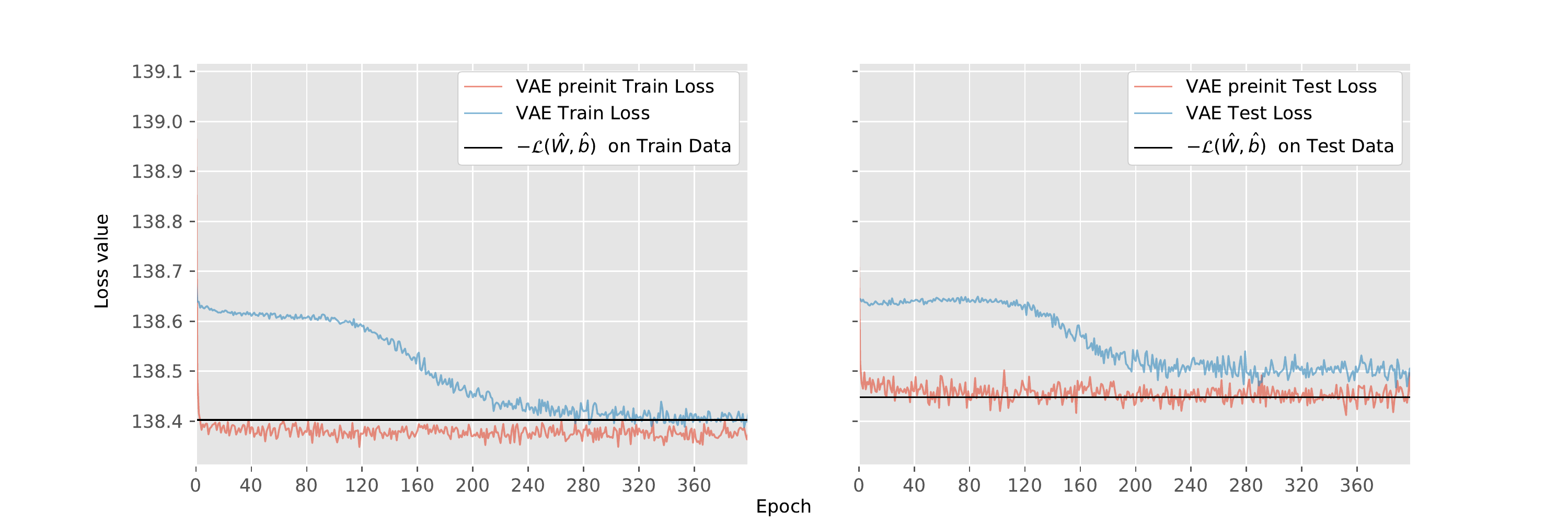}}
		\caption{The figure displays the behaviour of the loss of the simple VAEs as constructed in \ref{subsec:Loss development for ``canonical'' net architectures} over 400 epochs of training. The data is exactly as for figure \ref{fig:vae_vae_preinit_paper}. Further description can also be found there. }
		\label{fig:vae_vae_preinit_simple_paper}
	\end{center}
	\vskip -0.2in
\end{figure*}

For both nets, we set the ``canonical'' VAE architecture as follows:
\begin{align*}
x(d)\rightarrow E_1(2000)\rightarrow E_2(d)&\rightarrow \; \bmu_z(k)\; \rightarrow\;\hat{x}(d)\\
&\searrow\log\bsigma_z^2(k)\nearrow& \\
\end{align*}
This structure admits to the data generation in \ref{subsec:Data generation} and to our theoretical setting in \ref{subsec:Assumptions and lower bound}. \\
The decoder resembles exactly the assumed structure in proposition \ref{wohoo-i-got-a-proposition}. We still have several layers for the encoder in order to give the VAE the possibility to account for the assumptions in proposition \ref{wohoo-i-got-a-proposition}.

With this architecture, for the ``VAE preinit'' we do not have to consider not needed dimensions as zero (as mentioned in \ref{subsec:Initialization for Variational Autoencoder}) as there are none. 

Figure \ref{fig:vae_vae_preinit_simple_paper} shows the results for training ``canonical'' versions of VAEs. We see how the ``VAE preinit'' directly adapts the assumed bound for the training case, where the high values at the beginning originate from the not readily initialized layers $E_1$ and $E_2$ of the encoder. The same holds for the test case.\\
For the training case, the normal VAE (initialized as in \ref{subsec:VAE architecture}) takes a substantially longer time to adapt to the bound but eventually reaches it. For the test data case, we see that the normal VAE never performs at least as good as our lower bound indicates. \\
The ``VAE preinit'' is more favourable in both cases.

\end{document}